\documentclass[runningheads]{llncs}
\usepackage{graphicx}

\usepackage{tikz}
\usepackage{comment}
\usepackage{amsmath,amssymb} 
\usepackage{color}
\usepackage{bbold}
\usepackage{xcolor}
\usepackage{algpseudocode}
\usepackage{booktabs}
\usepackage{multirow}
\usepackage{amsmath}
\usepackage{pifont}
\usepackage{makecell}
\usepackage{diagbox}
\usepackage{graphicx}
\usepackage{colortbl}
\usepackage[symbol*]{footmisc}
\newcommand{\cmark}{\ding{51}}%
\newcommand{\xmark}{\ding{55}}%
\usepackage[accsupp]{axessibility}  
\usepackage[linesnumbered,ruled,vlined]{algorithm2e}
\SetKwInput{KwInput}{Input}                
\SetKwInput{KwOutput}{Initialize}              

\def \noisetype{\texttt{\textbf{\footnotesize{N-identities}$|$K$^C$-clusters}}}

\def \algname{Evolving Sub-centers Learning}
\def \nkc{$N$, $K$, and $C$}

\definecolor{c1}{RGB}{179,199,231}
\definecolor{c2}{RGB}{225,240,217}
\definecolor{c3}{RGB}{248,203,173}
\definecolor{c4}{RGB}{112,173,71}
\definecolor{c5}{RGB}{191,144,0}

\begin{document}
\pagestyle{headings}
\mainmatter
\def\ECCVSubNumber{1982}  

\title{Rethinking Robust Representation Learning Under Fine-grained Noisy Faces} 

\titlerunning{Rethinking Robust Representation Learning Under Fine-grained Noisy Faces}
%
\author{
Bingqi Ma\inst{1\ast} \and
Guanglu Song\inst{1\ast} \and
Boxiao Liu\inst{1,2} \and
Yu Liu\inst{1\dagger}
}

%
\authorrunning{B. Ma~et~al.}
%
\institute{Sensetime Research \and
SKLP, Institute of Computing Technology, CAS
\email{\{mabingqi,songguanglu\}@sensetime.com, liuboxiao@ict.ac.cn, liuyuisanai@gmail.com
}
}

\footnotetext[1]{Equal contributions.}
\footnotetext[2]{Corresponding author.}

\maketitle

\begin{abstract}
Learning robust feature representation from large-scale noisy faces stands out as one of the key challenges in high-performance face recognition.
Recent attempts have been made to cope with this challenge by alleviating the intra-class conflict
and inter-class conflict.
However, the unconstrained noise type in each conflict still makes it difficult for these algorithms to perform well.
To better understand this, we reformulate the noise type of each class in a more fine-grained manner as \noisetype{}. Different types of noisy faces can be generated by adjusting the values of \nkc.
Based on this unified formulation, we found that the main barrier behind the noise-robust representation learning is the flexibility of the algorithm under different \nkc.
For this potential problem, we propose a new method, named Evolving Sub-centers Learning~(ESL), to find optimal hyperplanes to accurately describe the latent space of massive noisy faces.
More specifically, we initialize $M$ sub-centers for each class and ESL encourages it to be automatically aligned to \noisetype{} faces via producing, merging, and dropping operations.
Images belonging to the same identity in noisy faces can effectively converge to the same sub-center and samples with different identities will be pushed away.
We inspect its effectiveness with an elaborate ablation study on the synthetic noisy dataset with different \nkc.
Without any bells and whistles, ESL can achieve significant performance gains over state-of-the-art methods on large-scale noisy faces.

\end{abstract}
\section{Introduction}
Owing to the rapid development of computer vision technology~\cite{song2020revisiting,song2018beyond,song2018region}, face recognition~\cite{wang2018cosface,deng2019arcface,liu2019towards,zhang2020discriminability,liu2018transductive} has made a remarkable improvement and has been widely applied in the industrial environment.
Much of this progress was sparked by the collection of large-scale web faces as well as the robust learning strategies~\cite{deng2019arcface,wang2018cosface} for representation learning. For instance, MS-Celeb-1M~(MS1MV0)~\cite{guo2016ms} provides more than 10 million face images with rough annotations.
The growing scale of training datasets inevitably introduces unconstrained noisy faces and can easily weaken the performance of state-of-the-art methods.
Learning robust feature representation from large-scale noisy faces has become an important challenge for high-performance face recognition.
Conventional noisy data learning, such as recursive clustering, cleaning, and training process, suffers from high computational complexity and cumulative error. 
For this problem, Sub-center ArcFace~\cite{deng2020sub} and SKH~\cite{liu2021switchable} are proposed to tackle the intra-class conflict or inter-class conflict by designing multiple sub-centers for each class.
These algorithms demonstrate remarkable performance in the specific manual noise. However, they are still susceptible to the unconstrained types of real-world noisy faces.
Naturally, we found that it is far from enough to just divide label noise in face recognition roughly into intra-class noise and inter-class noise. It greatly limits our understanding of the variant noise types and the exploration of noise-robust representation learning strategies.

\begin{figure}[t]
\centering
\includegraphics[width=\linewidth]{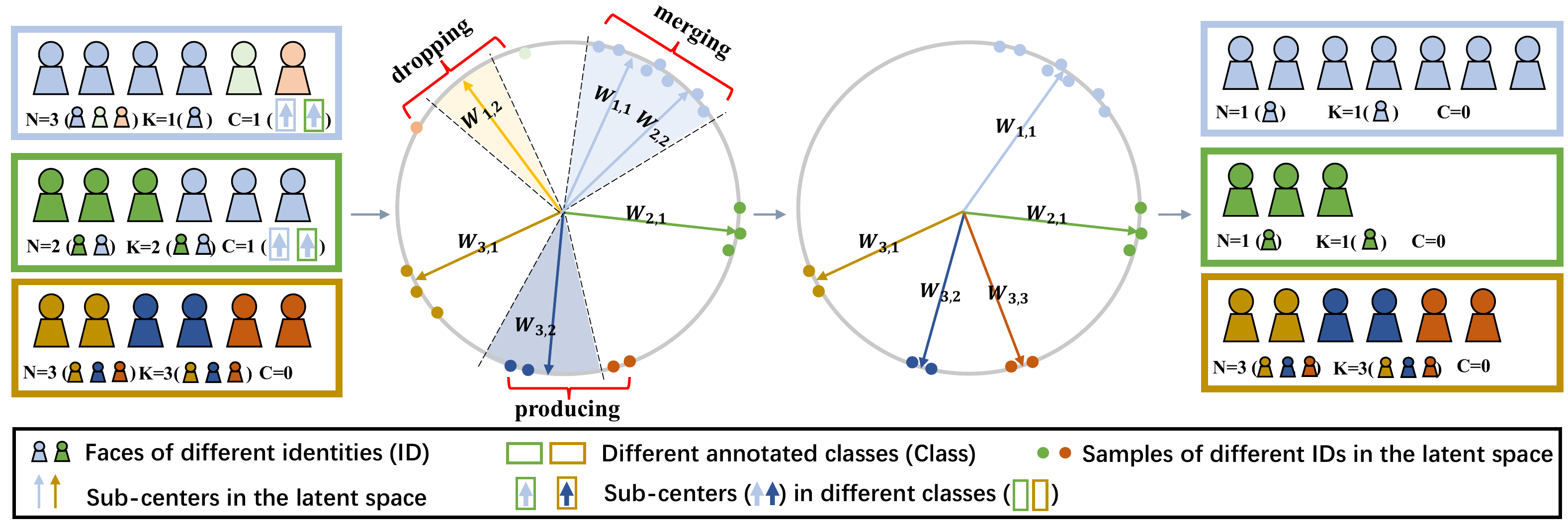}
\caption{Illustration of fine-grained noisy faces and ESL~(Best viewed in color). For \textcolor{c2}{ID$2$} and \textcolor{c3}{ID$3$} in \textcolor{c1}{Class1}, there is only one image for each ID, and they will be removed by the dropping operation. \textcolor{c1}{ID$1$} appears in both \textcolor{c1}{Class$1$} and \textcolor{c4}{Class$2$}, so that the merging operation will merge images in  \textcolor{c4}{Class$2$} with \textcolor{c1}{ID$1$} into  \textcolor{c1}{Class$1$}. In \textcolor{c5}{Class$3$}, there are $3$ IDs but only $2$ sub-centers, so the producing operation will produce another valid sub-center. Our proposed ESL can be flexibly adapted to different combinations of~$NKC$, and is more robust to unconstrained real-world noise.}
\label{fig:noise}
\end{figure}
To better understand this, we reformulate the noise data in a more fine-grained manner as \noisetype{} faces for each class.
Faces sharing \textit{identity}~(ID) means these images come from the same person.
Faces annotated with the same label construct a \textit{class}, and there may be annotation errors in the class.
If there are no less than two faces for an identity, these images build a meaningful \textit{cluster}~\cite{du2020semi}.
Please refer to the Sec.~1 in the appendix for the holistic description of terms and notations.
Taking the Class$1$ in Fig.~\ref{fig:noise} as an example, there are 3 IDs marked with ID$1$, ID$2$ and ID$3$, so the $N$ in Class$1$ is 3.
However, only ID$1$ contains more than 2 images, so the $K$ in Class$1$ is 1.
Furthermore, ID$1$ appears in both Class$1$ and Class$2$, which indicates one inter-class conflict, so the $C$ in Class$1$ is 1.
As shown in Table~\ref{tab:noise}, our proposed~\noisetype{} formulation can clearly represent different fine-grained noisy data. 
\begin{table}[]
    \centering
    \caption{Noise type in different combination of \nkc. 
    $\bigtriangleup$ represents intra-class conflict in which there are multiple clusters in the class. 
    $\Box$ represents intra-class conflict where there are outlier faces in the class. 
    $\Diamond$~represents the inter-class conflict in which there are multiple clusters with the same identify in different classes.} 
    \resizebox{.8\textwidth}{!}{
    \begin{tabular}{c|c|c|c|c|c}
    \toprule[1.5pt]
    & $N = K = 1$ & $N = K~\textgreater~1$ & $N~\textgreater~K~\textgreater~1$ & $N~\textgreater~K=1$ & $N~\textgreater~K=0$ \\
    \hline
    C = 0 & \textbf{-} & $\bigtriangleup$ & $\bigtriangleup$~$\square$ & $\square$ & $\square$ \\
    \hline
    C~\textgreater~0 & $\Diamond$ & $\bigtriangleup$~$\Diamond$ & $\bigtriangleup$~$\square$~$\Diamond$ & $\square$~$\Diamond$ & -\\
    \bottomrule[1.5pt]
    \end{tabular}
    }
    \label{tab:noise}
\end{table}

However, if $N$ and $K$ are larger than the predefined sub-center number in Sub-center ArcFace~\cite{deng2020sub} and SKH~\cite{liu2021switchable}, images without corresponding sub-center will lead to intra-class conflict. 
If $C$ exceeds the sub-center number in SKH~\cite{liu2021switchable}, extra conflicted clusters will bring inter-class conflict.
Both intra-class conflict and inter-class conflict will lead to the wrong gradient, which would dramatically impair the representation learning process.

In this paper, we constructively propose a flexible method, named Evolving Sub-centers Learning (ESL), to solve this problem caused by unconstrained \nkc.
More specifically, we initialize $M$ sub-centers for each class first.
Images belonging to the same identity will be pushed close to the corresponding positive sub-center and away from all other negative sub-centers.
Owning to elaborate designed producing, dropping, and merging operations, ESL encourages the number of sub-centers to be automatically aligned to \noisetype{} faces.
As shown in Fig.~\ref{fig:noise}, our proposed ESL can be flexibly adapted to different combinations of~$NKC$, and is more robust to unconstrained real-world noise. 
We inspect its effectiveness with elaborate ablation study on variant \noisetype{} faces. 
Without any bells and whistles, ESL can achieve significant performance gains over current state-of-the-art methods in large-scale noisy faces.
To sum up, the key contributions of this paper are as follows:

\begin{itemize}
  \item [-] 
  We reformulate the noise type of faces in each class into a more fine-grained manner as \noisetype{}.
  Based on this, we reveal that the key to robust representation learning strategies under real-world noise is the flexibility of the algorithm to the variation of \nkc. 
  \item [-]
  We introduce a general flexible method, named Evolving Sub-centers Learning (ESL), to improve the robustness of feature representation on noisy training data. The proposed ESL enjoys scalability to different combinations of \nkc, which is more robust to unconstrained real-world noise.
  \item [-]
  Without relying on any annotation post-processing or iterative training, ESL can easily achieve significant performance gains over state-of-the-art methods on large-scale noisy faces.
\end{itemize}
\section{Related work}
\textbf{Loss function for Face Recognition.} Deep face recognition models rely heavily on the loss function to learn discriminate feature representation. 
Previous works~\cite{deng2019arcface,sun2014deep,wang2018cosface,huang2020curricularface,schroff2015facenet,liu2017sphereface,wang2020mis,deng2017marginal,zhang2019adacos} usually leverage the margin penalty to optimize the intra-class distance and the inter-class distance. 
Facenet~\cite{schroff2015facenet} uses the Triplet to force that faces in different classes have a large Euclidean distance than faces in the same class. However, the Triplet loss can only optimize a subset of all classes in each iteration, which would lead to an under-fitting phenomenon. It is still a challenging task to enumerate the positive pairs and negative pairs with a growing number of training data. 
Compared with the sample-to-sample optimization strategy, Liu~et~al.~\cite{liu2017sphereface} proposes the angular softmax loss which enables convolutional neural networks to learn angularly discriminative features.
Wang~et~al.~\cite{wang2018cosface} reformulates Softmax-base loss into a cosine loss and introduces a cosine margin term to further maximize the decision margin in the angular space.
Deng~et~al.~\cite{deng2019arcface} directly introduces a fixed margin, maintaining the consistency of the margin in the angular space. 
Liu~et~al.~\cite{liu2019towards} adopts the hard example mining strategy to re-weight temperature in the Softmax-base loss function for more effective representation learning.\\
\textbf{Dataset for Face Recognition.} Large-scale training data can significantly improve the performance of face recognition models. MS1MV0~\cite{guo2016ms}, in which there are about $100$K identities and $1$M faces, is the most commonly used face recognition dataset. 
MS1MV3 is a cleaned version from MS1MV0 with a semi-automatic approach~\cite{deng2019lightweight}. 
An~et~al.~\cite{an2021partial} cleans and merges existing public face recognition datasets, then obtains Glint360K with $17$M faces and $360$K IDs.
Recently, Zhu~et~al.~\cite{zhu2021webface260m} proposes a large-scale face recognition dataset WebFace260M and a automatically cleaning pipeline.
By iterative training and cleaning, they proposed well-cleaned subset with $42$M images and $2$M IDs.\\
\textbf{Face Recognition under Noisy Data.}
Iterative training and cleaning is an effective data cleanup method. 
However, it is extremely inefficient as the face number increases. 
Recent works~\cite{wang2019co,hu2019noise,zhong2019unequal,deng2019arcface,liu2021switchable,zhu2004optimal,zhu2006subclass} focus on efficient noisy data cleanup methods.
Zhong~et~al.~\cite{zhong2019unequal} decouples head data and tail data of a long-tail distribution and designs a noise-robust loss function to learn the sample-to-center and sample-to-sample feature representation.
Deng~et~al.~\cite{deng2020sub} designs multiple centers for each class, splitting clean faces and noisy faces into different centers to deal with the inter-class noise.
Liu~et~al.~\cite{liu2021switchable} leverages multiple hyper-planes with a greedy switching mechanism to alleviate both inter-class noise and intra-class noise.
However, these methods are sensitive to hyper-parameter and can not tackle the complex noisy data distribution.
\section{The Proposed Approach}
\begin{figure}[t]
\centering
\includegraphics[width=\linewidth]{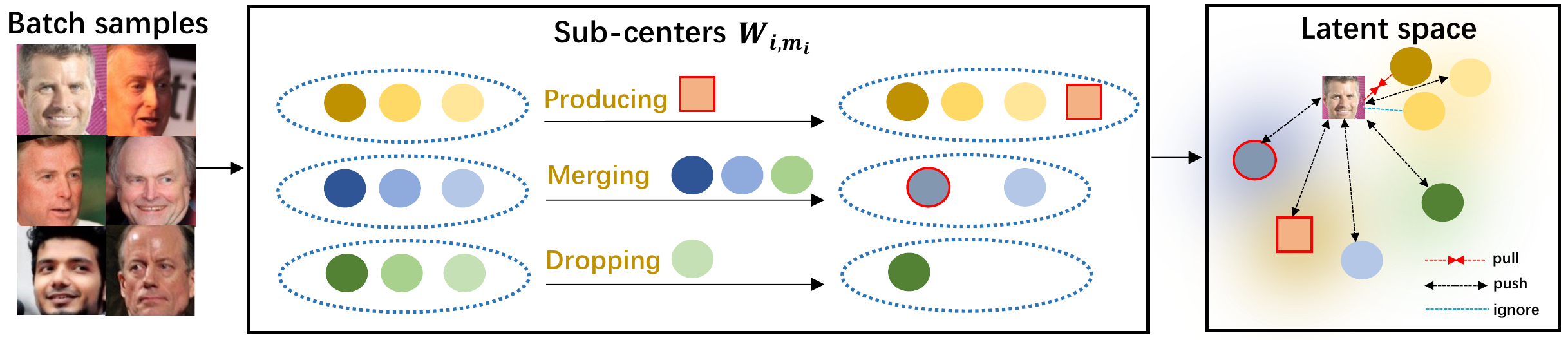}
\caption{The pipeline of the Evolving Sub-centers Learning. We initialize M sub-centers for each class and they will evolve adaptively to align the data distribution.
It pushes the images belonging to an identity close to the specific sub-center and away from all other negative sub-centers. The sub-center with a confusing similarity to the current sample will be ignored in the latent space. This can effectively dispose of the label conflict caused by fine-grained noisy faces.}
\label{fig:pipeline}
\end{figure}

In this section, we are committed to eliminating the unconstrained real-world noise via a flexible and scalable learning manner, named Evolving Sub-centers Learning, that can be easily plugged into any loss functions.
The pipeline of ESL is as shown in Fig.~\ref{fig:pipeline}.
We will first introduce our proposed ESL and then give a deep analysis to better understand its effectiveness and flexibility under fine-grained noisy faces. Finally, we conduct a detailed comparison between ESL and the current state-of-the-art noise-robust learning strategies.

\subsection{\algname{}}
In face recognition tasks, the unified loss function can be formulated as:
\begin{align}
  \mathcal{L}({x}_i) = -{\rm log}\frac{e^{\hat{f}_{i,y_i}}}{e^{\hat{f}_{i,y_i}} + \sum^{S}_{j=1,j\neq y_i}e^{f_{i,j}}}, \label{loss}
\end{align}
where $i$ is the index of face images, $y_i$ represents the label ID of image $I_i$ and
$S$ indicates the total class number in the training data.
Let the ${x}_i$ and ${W}_j$ denote the feature representation of face image $I_i$ and the $j$-th class center, 
the logits $\hat{f}_{i,y_i}$ and $f_{i,j}$ can be computed by:
\begin{align}
  \hat{f}_{i,y_i} &= s \cdot [m_1 \cdot {\rm cos}(\theta_{i, y_i}+m_2) -m_3], \\
  f_{i,j} &= s \cdot cos(\theta_{i,j}),
\end{align}
where $s$ is the re-scale parameter and $\theta_{i,j}$ is the angle between the ${x}_i$ and ${W}_j$ normalized with $\mathcal{L}_2$ manner.
For ArcFace with $m_1=1$ and $m_3=0$, we can compute the $\theta_{i,y_i}$ by:
\begin{align}
    \theta_{i,y_i} = arccos(\frac{W^T_{y_i}}{||W^T_{y_i}||_2}\frac{x_{i}}{||x_{i}||_2}).
\end{align}

As shown in SKH~\cite{liu2021switchable}, Eq.~(\ref{loss}) will easily get wrong loss under $N~\textgreater~1$ which indicates at least two different identities exist in the images currently labeled as the same identity.
In this paper, we address this problem by proposing the idea of using class-specific sub-centers for each class, which
can be directly adopted by any loss functions and will significantly increase its robustness.
As illustrated in Fig~\ref{fig:pipeline}, we init $M_j$ sub-centers for $j$-th class where each center is dominated by a learnable vector $W_{j, m_j}, m_j\in[1, M_j]$. The original class weight $W_j\in \mathbb{R}^{1\times D}$ can be replaced by all sub-centers $W_j\in \mathbb{R}^{1\times M_j\times D}$.
Based on this, Eq.~(\ref{loss}) can be re-written as:
 
\begin{align}
  \mathcal{L}({x}_i) = -{\rm log}\frac{e^{\hat{f}_{i,y_i, m}}}{e^{\hat{f}_{i,y_i, m}} + \sum^{j\in [1,C], m_j\in[1,M_j]}\limits_{(j, m_j) \neq (y_i, m)} (1-\mathbb{1}\{{\rm cos} (\theta_{i,j,m_j})~\textgreater~ \mathcal{D}_{j, m_j}\})e^{f_{i,j, m_j}}},   \label{subloss}
\end{align}
where indicator function $\mathbb{1}\{{\rm cos} (\theta_{i,j,m_j})~\textgreater~\mathcal{D}_{j, m_j}\}$ returns 1 when ${\rm cos} (\theta_{i,j,m_j})~\textgreater~\mathcal{D}_{j, m_j}$ and 0 otherwise.
We calculate the mean $\mu_{j, m_j}$ and standard deviation $\sigma_{j, m_j}$ of the cosine similarity between the sub-center $W_{j,m_j}$ and the samples belonging to it. 
$\mathcal{D}_{j, m_j}$ can be generated by:
\begin{align}
    \mathcal{D}_{j, m_j} = \mu_{j, m_j} + \lambda_1\sigma_{j, m_j}
\end{align}
$\hat{f}_{i,y_i, m}$ and $f_{i,j, m_j}$ are computed by:
\begin{align}
  \hat{f}_{i,y_i, m} &= s \cdot [m_1 \cdot {\rm cos}(\theta_{i, y_i, m}+m_2) -m_3], \\
  f_{i,j, m_j} &= s \cdot cos(\theta_{i,j, m_j}),
\end{align}
where $\theta_{i, j, m_j}$ is the angle between the feature representation $x_i$ of the $i$-th face image and the $m_j$-th sub-center $W_{j,m_j}$ in $j$-th class.
We determine $m$ for sample $I_i$ by the nearest distance priority manner as:
\begin{align}
	m = \mathop{\arg\max}_{m_{y_i}}  \mathrm{cos(\theta_{i, y_i, m_{y_i}})}. \ \ m_{y_i} \in [1, M_{y_i}] \label{m_com}
\end{align}

\begin{figure}[t]
\centering
\includegraphics[width=\linewidth]{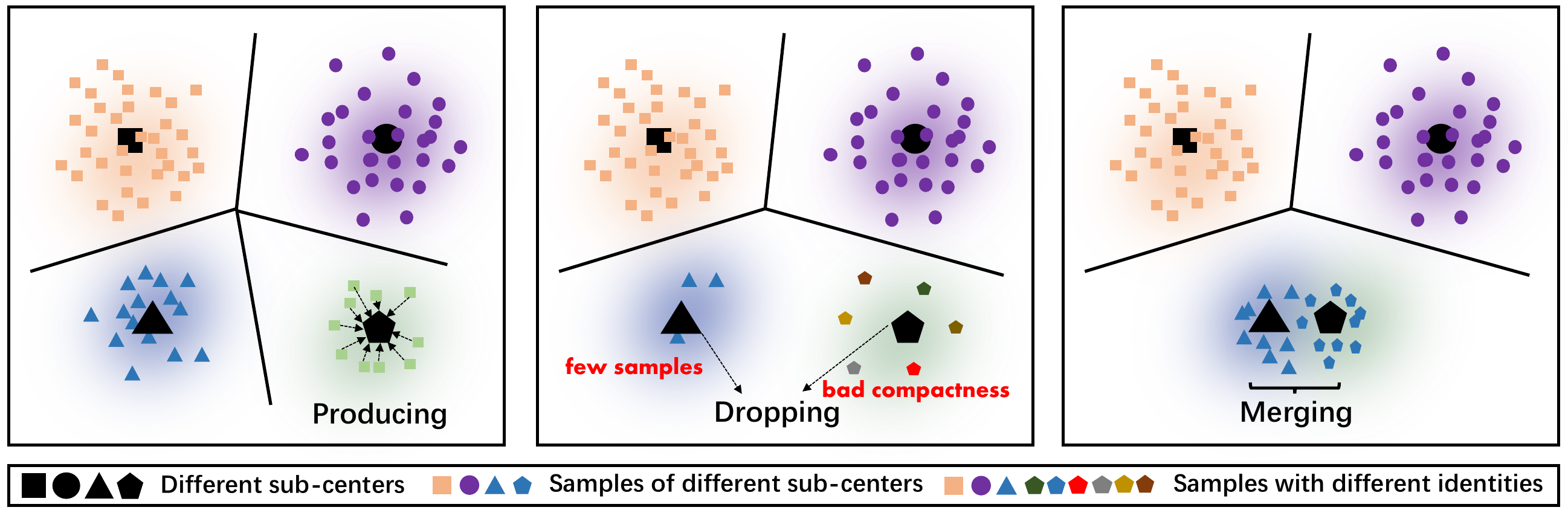}
\caption{Illustration of the sub-centers producing, dropping and merging. The instances with black color indicate the sub-centers. The instance belonging to each sub-center is represented by the same shape and different colors mean different identities.}
\label{fig:operation}
\end{figure}

Given an initial $M_j$ for each class, Eq.~(\ref{subloss}) can capture the unconstrained distribution of the whole training data with potential label noise.
It pushes the images belonging to the same identity close to a specific sub-center and away from all other negative sub-centers.
Meanwhile, the sub-center with a confusing similarity to the current sample will be ignored to dispose of the label conflict.
To make it more flexible with unconstrained changes in $N$, $K$, and $C$, we further introduce the producing, merging, and dropping operations as shown in Fig.~\ref{fig:operation}. \\
\textbf{Sub-centers Producing.} Based on the aforementioned design, it can effectively alleviate the conflict caused by label noise when $M_{j}~\textgreater~N~\textgreater~1$.
However, the unconstrained N makes it difficult to select the appropriate $M_{j}$ for each class.
To make it more flexible, we introduce the sub-centers producing operation to automatically align the sub-centers and the actual identity number in each class.
Given $\mathcal{N}$ images with label $y$ and assigned to sub-center $m$ with Eq.~\ref{m_com}, a new sub-center $W_{y,M_y+m}$ can be generated by:
\begin{align}
	W_{y,M_y+m} = \frac{1}{\mathcal{T}} \sum^{\mathcal{N}}_{i=1}\mathbb{1}\{cos(\theta_{i,y,m}) \textless \mu_{y,m} - \lambda_2 \sigma_{y,m}\}x_i, \ \ if \ \ \mathcal{T}~\textgreater~0, \label{producing}
\end{align}
where $\mathcal{T} = \sum^\mathcal{N}_{i=1} \mathbb{1}\{cos(\theta_{i,y,m}) \textless \mu_{y,m} - \lambda_2 \sigma_{y,m}\}$.
If $\mathcal{T} = 0$, there is no new sub-center to be formed.
After this, we can progressively produce new sub-center to house additional identities beyond $M_j$.
It effectively improves intra-class compactness and reduces the conflict caused by unconstrained $N$ and $K$.\\
\textbf{Sub-centers Dropping.} 
As demonstrated by~\cite{deng2020sub,liu2021switchable}, many state-of-the-art methods are susceptible to the outlier faces (the image number belonging to an identity is less than $2$, $N~\textgreater~K \geq 1$). These outlier images are hard to be pushed close to any corresponding positive sub-center. During the producing process, outlier images from each sub-center will generate a new sub-center. The dropping operation should remove the sub-center from outlier images but preserve the sub-center with a valid identity. Considering the standard deviation can not reflect the density of a distribution, we just leverage $\mu_{i,m_i}$ as the metric. The condition of dropping can be formulated as :
\begin{align}
    \mathcal{J}(W_{i, m_i}) = \mathbb{1}\{\mu_{i,m_i} \leq \lambda_3\}.
    \label{dropping}
\end{align}
If the $\mu_{i,m_i}$ is less than $\lambda_3$, we will ignore these images during the training process and then erase the specific sub-center.
\\
\textbf{Sub-centers Merging.}
Using sub-centers for each class can dramatically improve the robustness under noise.
However, the inter-class discrepancy will be inevitably affected by inter-class conflict caused by the shared identity between different sub-centers.
SKH~\cite{liu2021switchable} sets the same fixed number of sub-centers for each class, which can not handle complex inter-class conflict with unconstrained $C$.
Meanwhile, the sub-center strategy undermines the intra-class compactness as the samples in a clean class also converge to different sub-centers. 
To deal with this potential problem, we employ the sub-centers merging operation to 
aggregate different $W_{*,*}$. 
The condition of merging can be formulated as:
\begin{align}
	\mathcal{J}(W_{i,m_i}, W_{j, m_j}) = \mathbb{1}\{W^T_{i,m_i}W_{j, m_j} \geq {\rm max}(
	\mu_{j,m_j} + \lambda_4 \sigma_{j,m_j}, \mu_{i,m_i} + \lambda_4 \sigma_{i,m_i})\}, \label{merge}
\end{align}
where $W_{i,m_i}$ and $W_{j, m_j}$ are normalized with $L_2$ manner.
According to Eq.~(\ref{merge}), we merge multiple sub-centers satisfying $\mathcal{J}(*,*) = 1$ into a group and combine them into a single sub-center as following:
\begin{align}
    W_{*,new}=\frac{1}{|G|} \sum_{(p,m_p)\in G}W_{p,m_p},
    \label{merge_new}
\end{align}
where $G$ and $|G|$ indicate the merged group and its sub-center number.
Furthermore, images belonging to $G$ will be assigned to a new label (we directly select the minimum label ID in the $G$ as the target).
\subsection{Progressive Training Framework}
At the training stage, we perform the sub-centers producing, dropping and merging operations progressively to effectively alleviate the label conflict cause by unconstrained \nkc.
The training framework is summarized in Alg.~\ref{aug1}.
\resizebox{.9\textwidth}{!}{
\begin{minipage}{\linewidth}{
\begin{algorithm}[H]{
\DontPrintSemicolon
\caption{Evolving Sub-centers Learning}
\label{aug1}
    \KwInput{Training data set $\mathcal{X}$, label set $\mathcal{Y}$, total training epoch E, start epoch $\varepsilon$ for ESL.}
    \KwOutput{Label number $C$, sub-centers number $M_{*}$ and $W_{*,*}$ for each class.}
    \begin{algorithmic}
    \State $e\gets 0$; \\   
    \While{e \textless~E}{
   	    \State sample data $X$, $Y$ from $\mathcal{X}$, $\mathcal{Y}$;
   		\State compute loss function $L(X,Y)$ by Eq.~(\ref{subloss}) and update model;
   		\State generating $\mu_*$ and $\sigma_*$ for each sub-center;\\
   		\If{e~\textgreater~$\varepsilon$}
   		{ 
   		    \For{$i=1$ \KwTo $C$}{
   		        \For{$j=1$ \KwTo $M_{i}$}{
   		            \textcolor{blue}{\tcp*[l]{Producing}}
   		            \State computing $\mathcal{T}$ via Eq.~(\ref{producing});\\
                    \If{$\mathcal{T} > 0$}{
                        \State computing $W_{i,M_i+j}$ via Eq.~(\ref{producing});
                    }
                    \textcolor{blue}{\tcp*[l]{Dropping}}
                    \State generating $\mathcal{J}(W_{i, j})$ via Eq.~(\ref{dropping});\\
   		            \If{$\mathcal{J}(W_{i, j})$}{
                        \State dropping sub-center $W_{i, j}$;
                        \State generating $\mathcal{X}_i$ as images with label $i$;\\
   		                \ForEach{image $X$ in $\mathcal{X}_i$}{
   		                    \State computing $m$ via Eq.(~\ref{m_com});\\
   		                    \If{j = m}{
   		                        \State dropping image $X$;
   		                    }
   		                }
   		            }
   		        }
   		    }
   		    \textcolor{blue}{\tcp*[l]{Merging}}
   		    \State generating vertex set $V$ with each sub-center $W$ in $W_{*,*}$; 
   		    \State generating edge set $E$ with $(W_{i}, W_{j})$ if $\mathcal{J}(W_{i}, W_{j})=1$ via Eq.~(\ref{merge}); 
   		    \State generating graph $G=$($V$, $E$); \\
            \ForEach{connected component $g$ in $G$}{
                \State generating new sub-center via Eq.~(\ref{merge_new}); 
                \State relabeling images belonging to sub-centers in $g$;
            }
   		}
   		$e\gets e+1$; 
    }
    \end{algorithmic}
}
\end{algorithm}
}
\end{minipage}
} \\
In this manner, ESL is able to capture the complex distribution of the whole training data with unconstrained label noise.
It tends to automatically adjust the sub-centers to align the distribution of \nkc~in the given datasets. 
This allows it to be flexible in solving the real-world noise while preventing the network from damaging the inter-class discrepancy on clean faces.
\subsection{Robustness Analysis on Fine-grained Noisy Faces}
When applied to the practical \noisetype{} faces, the key challenge is to process different combinations of \nkc. 
In Tab.~\ref{tab:noise}, we have analyzed the noise type in different combinations of \nkc. Now we investigate the robustness of ESL on fine-grained noisy faces.

To simplify the analysis, we first only consider the circumstance when $\mathbf{C=0}$. 
(1) $\mathbf{N=K=1,C=0}$. This phenomenon indicates the training dataset is absolutely clean. In this manner, most of the feature learning strategies can perform excellent accuracy. However, introducing the sub-centers for each class will damage the intra-class compactness and degrade the performance. 
The sub-centers merging allows ESL to progressively aggregate the sub-centers via Eq.~(\ref{merge}) to maintain the intra-class compactness.
(2) $\mathbf{N=K~\textgreater~1,C=0}$. This means there are several identities existing in a specific class. $N=K$ represents the images for each identity are enough to form a valid cluster in the latent space and there are no outlier images.
Under this manner, with appropriate hyper-parameter, the state-of-the-art methods Sub-center ArcFace~\cite{deng2020sub} and SKH~\cite{liu2021switchable} can effectively cope with this label conflict. 
However, the unconstrained $N$ and $K$ still make them ineffective even with some performance gains.
In ESL, the sub-centers producing via Eq.~(\ref{producing}) adaptively produce new sub-centers to accommodate the external identities beyond the initialized sub-center number if there are fewer sub-centers than identities in the class. 
If the number of sub-centers is larger than the identity number, the merging strategy will merge clusters with the same identity to keep the intra-class compactness.
(3) $\mathbf{N~\textgreater~K~\textgreater~1,C=0}$. Besides the conflict clusters, several identities can not converge to valid clusters in the latent space. We find that this is caused by the few-shot samples in each identity.
It lacks intra-class diversity, which prevents the network from effective optimization and leads to the collapse of the feature dimension. 
To deal with these indiscoverable outliers, we design the sub-centers dropping operation to discard these sub-centers with few samples or slack intra-class compactness based on Eq.~\ref{dropping} in ESL. This is based on our observation that these sub-centers are not dominated by any one identity. Multiple outliers try to compete for the dominance, leading to bad compactness.
(4) $\mathbf{N~\textgreater~K=1,C=0}$. It indicates that there is one valid identity and several outlier images in this class. ESL will enable the dropping strategy to remove the noise images and keep valid faces. 
(5) $\mathbf{N~\textgreater~K=0,C=0}$. It indicates each identity in the class only owns few-shot samples. We proposed dropping operation will discard all the sub-centers in this class. 
For $\mathbf{C~\textgreater~0}$, there are multiple clusters with the same identity but different labels. This introduces the inter-class conflict. 
The state-of-the-art method SKH~\cite{liu2021switchable} can not perform well under unconstrained $C$.
For this potential conflict, Eq.~(\ref{merge}) in ESL can also accurately alleviate this by dynamically adjusting the label of images belonging to the merged sub-centers.

\subsection{Comparison with Other Noise-robust Learning Strategies}
The main difference between the proposed ESL and other methods~\cite{deng2020sub,liu2021switchable,hu2019noise,zhong2019unequal,deng2019arcface}
is that ESL is less affected by the unconstrained \nkc~from the real-world noise. 
It's more flexible to face recognition under different types of noise while keeping extreme simplicity, only adding three sub-centers operation.
To better demonstrate this, we make a detailed comparison with other methods under fine-grained noisy faces as shown in Tab.~\ref{tab:headings}.
The superiority of our method is mainly due to the flexible sub-center evolving strategy, which can handle variant intra-class noise and inter-class noise simultaneously.
\begin{table}[t]
    \centering
    \caption{Comparison with other noise-robust learning strategies under different types of noise. $+$ indicates the method can solve the noise under the specific setting. $+$$+$$+$ indicates the method can solve the noise problem. $-$ indicates the method can not handle the problem.}
    \resizebox{.8\textwidth}{!}{
    \begin{tabular}{l|c|c|c|c|c}
    \toprule[1.5pt]
          Method & C & $N = K = 1$ & $N = K~\textgreater~1$ & $N~\textgreater~K~\textgreater~1$ & $N~\textgreater~1 \geq K$ \\
           \hline
           \multirow{2}{*}{ArcFace~\cite{deng2019arcface}} & $C = 0$ & +++ & -  & - & -  \\
           \cline{2-6}
           & $C~\textgreater~0$ & - & - & - & - \\
           \hline
           \multirow{2}{*}{NT~\cite{hu2019noise}} & $C = 0$ & +++ & - & + & + \\
           \cline{2-6}
           & $C~\textgreater~0$ & - & - & - & - \\
           \hline
           \multirow{2}{*}{NR~\cite{zhong2019unequal}} & $C = 0$ & +++ & - & + & + \\
           \cline{2-6}
           & $C~\textgreater~0$ & - & - & - & - \\
           \hline
           \multirow{2}{*}{Sub-center~\cite{deng2020sub}} & $C = 0$ & + & + & + & + \\
           \cline{2-6}
           & $C~\textgreater~0$ & -  & - & - & - \\
           \hline
           \multirow{2}{*}{SKH~\cite{liu2021switchable}} & $C = 0$ & + & + & + & + \\
           \cline{2-6}
           & $C~\textgreater~0$ & + & + & + & + \\
           \hline
           \multirow{2}{*}{ESL} & $C = 0$ & +++ & +++ & +++ & +++ \\
           \cline{2-6}
           & $C~\textgreater~0$ & +++ & +++ & +++ & +++ \\
    \bottomrule[1.5pt]
    \end{tabular}
    }
    \label{tab:headings}
\end{table}
\section{Experiments}
\subsection{Experimental Settings}
\textbf{Datasets.}
MS1MV0~\cite{guo2016ms} and MS1MV3~\cite{deng2019lightweight} are popular academic face recognition datasets.
MS1MV0~\cite{guo2016ms} is raw data that is collected from the search engine based on a name list, in which there is around $50\%$ noise.
MS1MV3~\cite{deng2019lightweight} is the cleaned version of MS1MV0~\cite{guo2016ms} by a semi-automatic pipeline.
To further explore the effectiveness of our proposed ESL, we also elaborately construct synthetic noisy datasets. We establish intra-conflict, inter-class conflict, and mixture conflict noisy datasets, which will be detailed introduced in the supplementary material. 
As for the performance evaluation, we tackle the True Accept Rate~(TAR) at a specific False Accept Rate~(FAR) as the metric.
We mainly consider the performance on IJB-B~\cite{whitelam2017iarpa} dataset and IJB-C~\cite{maze2018iarpa} dataset. 
Moreover, we also report the results on LFW~\cite{huang2008labeled}, CFP-FP~\cite{sengupta2016frontal} and AgeDB-30~\cite{moschoglou2017agedb}.
\label{synthetic}
\\
\textbf{Implementation Details.}
Following ArcFace~\cite{deng2019arcface}, we generate aligned faces with RetinaFace~\cite{deng2020retinaface} and resize images to $(112\times112)$.
We employ ResNet-50~\cite{he2016deep} as backbone network to extract $512$-D feature embedding.
For the experiments in our paper, we initialize the learning rate with $0.1$ and divide it by $10$ at $100$K, $160$K, and $220$K iteration. 
The total training iteration number is set as $240$K. 
We adopt an SGD optimizer, then set momentum as $0.9$ and weight decay as $5e$-$4$. 
The model is trained on $8$ NVIDIA A$100$ GPUs with a total batch size of $512$.
The experiments are implemented with Pytorch~\cite{paszke2019pytorch} framework.
For experiments on ESL, we set the initial number of sub-centers for each class as $3$.
The $\lambda_1$, $\lambda_2$, $\lambda_3$ and $\lambda_4$ is separately set as $2$, $2$, $0.25$, and $3$.
\subsection{Comparison with State-of-the-art}
We conduct extensive experiments to investigate our proposed \algname{}. 
In Table.~\ref{main_exp}, we compare ESL with state-of-the-art methods on both real-word noisy dataset MS1MV0~\cite{guo2016ms} and the synthetic mixture of noise dataset. Without special instructions, the noise ratio is $50\%$. 
\begin{table}[t]
    \centering
    \caption{Experiments of different settings on MS1MV0 and synthetic mixture noisy dataset comparing with state-of-the-art methods.}
    \resizebox{\textwidth}{!}{
    \begin{tabular}{l|c|c c c|c c c}
    \toprule[1.5pt]
        \multirow{2}{*}{Method} & \multirow{2}{*}{Dataset} &  & IJB-B &  &  & IJB-C & \\
                &         & $1e$$-$$3$ & $\mathbf{1e}$$\mathbf{-}$$\mathbf{4}$ & $1e$$-$$5$ & $1e$$-$$3$ & $\mathbf{1e}$$\mathbf{-}$$\mathbf{4}$ & $1e$$-$$5$  \\
        \hline
        ArcFace~\cite{deng2019arcface}                   & MS1MV0 & 93.27 & 87.87 & 74.74 & 94.59 & 90.27 & 81.11 \\ 
        Sub-center ArcFace M=3~\cite{deng2020sub}    & MS1MV0 & 94.88 & 91.70 & 85.62 & 95.98 & 93.72 & 90.59 \\
        Co-ming~\cite{wang2019co} & MS1MV0 & 94.99 & 91.80 & 85.57 & 95.95 & 93.82 & 90.71 \\
        NT~\cite{hu2019noise}     & MS1MV0 & 94.79 & 91.57 & 85.56 & 95.86 & 93.65 & 90.48 \\
        NR~\cite{zhong2019unequal}& MS1MV0 & 94.77 & 91.58 & 85.53 & 95.88 & 93.60 & 90.41 \\
        SKH + ArcFace M=3~\cite{liu2021switchable}         & MS1MV0 & 95.89 & 93.50 & 89.34 & 96.85 & 95.25 & 93.00 \\
        \hline
        ESL + ArcFace             & MS1MV0 & $\mathbf{96.61}$ & $\mathbf{94.60}$ & $\mathbf{91.15}$ & $\mathbf{97.58}$ & $\mathbf{96.23}$ & $\mathbf{94.24}$ \\
        \hline
        ArcFace~\cite{deng2019arcface}                   & Mixture of Noises & 93.17 & 87.54 & 74.02 & 94.99 & 90.03 & 82.40 \\
        Sub-center ArcFace M=3~\cite{deng2020sub}    & Mixture of Noises & 92.83 & 86.80 & 73.11 & 94.20 & 89.32 & 81.43 \\
        SKH + ArcFace M=4~\cite{liu2021switchable}         & Mixture of Noises & 95.76 & 93.62 & 89.18 & 96.89 & 95.16 & 92.71 \\
        \hline
        ESL + ArcFace             & Mixture of Noises & $\mathbf{96.48}$ & $\mathbf{94.51}$ & $\mathbf{90.95}$ & $\mathbf{97.62}$ & $\mathbf{96.22}$ & $\mathbf{93.60}$ \\
    \bottomrule[1.5pt]
    \end{tabular}
    }
    \label{main_exp}
\end{table}

When training on noisy data, ArcFace has an obvious performance drop. It demonstrates that noise samples would do dramatically harm to the optimization process.
ESL can easily outperform current methods by an obvious margin. To be specific, ESL can outperform Sub-center ArcFace~\cite{deng2020sub} by $2.51\%$ and SKH~\cite{liu2021switchable} by $0.98\%$ on IJB-C dataset.
On the synthetic mixture noisy dataset, we make a grid search of the sub-center number in Sub-center ArcFace~\cite{deng2020sub} and SKH~\cite{liu2021switchable}. 
Sub-center ArcFace~\cite{deng2020sub} achieves the best performance when M$=$3 and SKH~\cite{liu2021switchable} achieves the best performance when M$=$4.
ESL can easily outperform Sub-center ArcFace~\cite{deng2020sub} by $6.9\%$ and SKH~\cite{liu2021switchable} by $1.06\%$ on IJB-C dataset.
Our proposed ESL can handle the fine-grained intra-class conflict and inter-class conflict under unconstrained \nkc, which brings significant performance improvement.

\subsection{Ablation Study}
\textbf{Exploration on Hyperparameters.}
The hyperparameters in our proposed ESL contain the initial sub-center number for each class and the $\lambda$ in each proposed operation. In Tab.~\ref{ablation_hyperparameter}, we investigate the impact of each hyperparameter.
\begin{table}[t]
    \centering
    \caption{Ablation experiments to explore the hyperparameters. }
    \resizebox{.8\textwidth}{!}{
    \begin{tabular}{ c | c | c | c | c | c}
    \toprule[1.5pt]
     $\lambda_1$(Eq.~(\ref{subloss})) & $\lambda_2$(Eq.~(\ref{producing})) & $\lambda_3$(Eq.~(\ref{dropping}))& $\lambda_4$(Eq.~(\ref{merge}))& $M_j$(Eq.~(\ref{subloss}))& TAR@FAR=-4 \\
     \hline
     \rowcolor{gray!15} 2 & 2 & 0.25 & 3 & 3 & $\mathbf{96.22}$ \\
     \hline
     1 & 2 & 0.25 & 3 & 3 & 95.73 \\
     3 & 2 & 0.25 & 3 & 3 & 95.88 \\
     \hline
     2 & 1 & 0.25 & 3 & 3 & 96.11 \\
     2 & 3 & 0.25 & 3 & 3 & 95.89 \\
     \hline
     2 & 2 & 0.2 & 3 & 3 & 96.05 \\
     2 & 2 & 0.3 & 3 & 3 & 96.18 \\
     \hline
     2 & 2 & 0.25 & 1 & 3 & 95.07 \\
     2 & 2 & 0.25 & 2 & 3 & 95.82 \\
     \hline
     2 & 2 & 0.25 & 3 & 1 & 96.03 \\
     2 & 2 & 0.25 & 3 & 2 & 96.14 \\
     2 & 2 & 0.25 & 3 & 4 & 96.20 \\
     2 & 2 & 0.25 & 3 & 5 & 96.19 \\
    \bottomrule[1.5pt]
    \end{tabular}
    }
    \label{ablation_hyperparameter}
\end{table}
\\
\textbf{Effectiveness of Proposed Operations.}
To demonstrate the effectiveness of our proposed ESL, we decouple each operation to ablate each of them on the mixture noisy dataset in Tab.~\ref{ablation}.
\begin{table}[t]
    \centering
    \caption{Ablation experiments to verify the effectiveness of proposed operations.}
    \resizebox{.8\textwidth}{!}{
    \begin{tabular}{l |c | c | c | c | c | c c c}
    \toprule[1.5pt]
    \multirow{2}{*}{ArcFace} & class-specific & \multirow{2}{*}{Merging} & \multirow{2}{*}{Producing} & \multirow{2}{*}{Dropping} & \multirow{2}{*}{Dataset} & & IJB-C & \\
     &  sub-center &  & & & &  $1e$$-$$3$ & $\mathbf{1e}$$\mathbf{-}$$\mathbf{4}$ & $1e$$-$$5$ \\
     \hline
     \makecell[c]{\cmark}& \makecell[c]{\xmark} & \makecell[c]{\xmark} & \makecell[c]{\xmark} & \makecell[c]{\xmark} & Mixture of Noises & 94.99 & 90.03 & 82.40 \\
     \makecell[c]{\cmark} & \makecell[c]{\cmark} & \makecell[c]{\xmark} & \makecell[c]{\xmark} & \makecell[c]{\xmark} & Mixture of Noises & 95.35 & 93.76 & 90.88 \\
     \makecell[c]{\cmark} & \makecell[c]{\cmark} & \makecell[c]{\cmark} & \makecell[c]{\xmark} & \makecell[c]{\xmark} & Mixture of Noises & 96.02 & 94.51 & 92.14 \\
     \makecell[c]{\cmark} & \makecell[c]{\cmark} & \makecell[c]{\cmark} & \makecell[c]{\cmark} & \makecell[c]{\xmark} & Mixture of Noises & 97.23 & 95.54 & 92.98 \\
     \makecell[c]{\cmark} & \makecell[c]{\cmark} & \makecell[c]{\cmark} & \makecell[c]{\cmark} & \makecell[c]{\cmark} & Mixture of Noises & \textbf{97.62} & \textbf{96.22} & \textbf{93.60} \\
    \bottomrule[1.5pt]
    \end{tabular}
    }
    \label{ablation}
\end{table}

\begin{figure}[h]
    \centering
    \includegraphics[width=\linewidth]{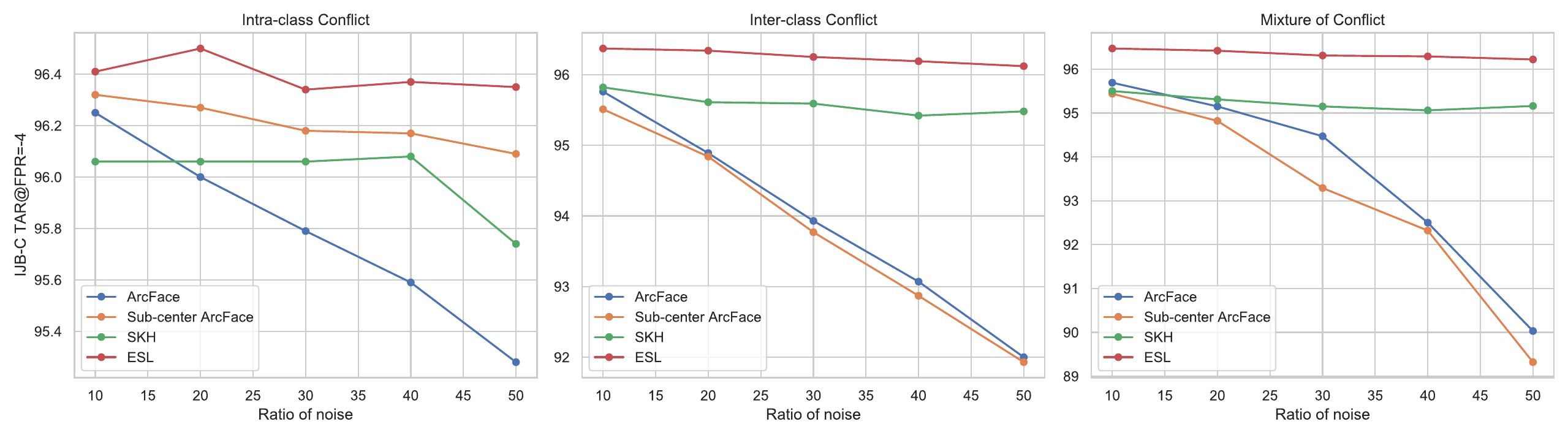}
    \caption{Experiments of ArcFace, Sub-center ArcFace, SKH and ESL on different noise ratio. We tackle the TAR@FAR$=$-$4$ on IJB-C dataset as evaluation metric.}
    \label{fig:noise_ratio}
\end{figure}

We take turns adding each component to the original ArcFace~\cite{deng2019arcface} baseline.
Due to the gradient conflict from massive fine-grained intra-class noise and inter-class noise, ArcFace~\cite{deng2019arcface} only achieves limited performance.
Sub-center loss introduces sub-centers for each identity to deal with the intra-class conflict. 
Meanwhile, the ignore strategy can ease part of conflict from inter-class noise. 
Sub-center loss brings $3.73\%$ performance improvement.
The merging operation aims to merge images that share the same identity but belong to different sub-centers.
The merging operation boosts the performance by $0.75\%$.
The producing operation can automatically align the sub-centers and the actual identity number in each class, which improves intra-class compactness effectively.
It further brings $1.03\%$ performance improvement.
The dropping operation tends to drop the outlier faces without the specific positive sub-center. These faces are hard to optimize and would harm the optimization process.
It can obtain a significant performance gain by $0.68\%$ under the fine-grained noisy dataset.\\
\textbf{Efficiency of ESL.}
Deng~et~al.~\cite{deng2020sub} and Liu~et~al.~\cite{liu2021switchable} adopt a posterior data clean strategy to filter out noise samples in an offline manner.
Deng~et~al.~\cite{deng2020sub} searches for the intra-class margin to drop the outlier samples for each domain center. Liu~et~al.\cite{liu2021switchable} further introduces inter-class margin to merge samples belonging to different centers. For each margin setting, they should train for $20$ epochs to verify its effectiveness, which is extremely time and computation resources consuming. In Table.~\ref{efficiency}, we compare ESL with these posterior cleaning strategies. ESL can also achieve better performance on both MS1MV0~\cite{guo2016ms} and synthetic mixture noisy dataset. Meanwhile, there is only a slight gap between ESL and ArcFace~\cite{deng2019arcface} training on cleaned MS1MV3.
\begin{table}[t]
    \centering
    \caption{Ablation experiments to compare ESL with posterior cleaning methods. The GPU hour is measured on NVIDIA A100 GPU. M=$n\downarrow1$ indicates the posterior data clean strategy proposed in Sub-center Arcface~\cite{deng2020sub}.}
    \resizebox{.8\textwidth}{!}{
    \begin{tabular}{l|c|c|c|c c c}
    \toprule[1.5pt]
        \multirow{2}{*}{Method} & \multirow{2}{*}{Dataset} &  \multirow{2}{*}{Posterior Clean} & \multirow{2}{*}{GPU Hour} &  & IJB-C & \\
                &     & &    & $1e$$-$$3$ & $\mathbf{1e}$$\mathbf{-}$$\mathbf{4}$ & $1e$$-$$5$   \\
        \hline
        Sub-center ArcFace M=$3\downarrow1$ & MS1MV0 & \makecell[c]{\cmark} & $128$ &  97.40 & 95.92 & 94.03 \\
        SKH + ArcFace M=$3\downarrow1$ & MS1MV0 & \makecell[c]{\cmark} & $128$ & 96.55 & 96.26 & 94.18 \\     
        ESL + ArcFace             & MS1MV0 & \makecell[c]{\xmark} & $\mathbf{80}$ & $\mathbf{97.58}$ & 96.23 & $\mathbf{94.24}$ \\
        \hline
        ArcFace & MS1MV3 & \makecell[c]{\xmark} &  $64$ & 97.64  & 96.44 & 94.66  \\
        \hline
        Sub-center ArcFace M=$3\downarrow1$   & Mixture of Noises & \makecell[c]{\cmark} & $128$ & 97.13 & 95.89 & 92.67 \\
        SKH + ArcFace M=$4\downarrow1$        & Mixture of Noises & \makecell[c]{\cmark} & $128$ & 97.46 & 96.14 & 92.87 \\
        ESL + ArcFace             & Mixture of Noises & \makecell[c]{\xmark} & $\mathbf{80}$ &  $\mathbf{97.62}$ & $\mathbf{96.22}$ & $\mathbf{93.60}$ \\
    \bottomrule[1.5pt]
    \end{tabular}
    }
    \label{efficiency}
\end{table}
\\
\textbf{Robustness under Various Noise Ratio.}
To further investigate the effectiveness of our proposed ESL, we conduct sufficient experiments under various noise ratios. As shown in Fig.~\ref{fig:noise_ratio}, we visualize the relationship between noise ratio and evaluation results. ESL can remain robustness under different noise ratio and surpass Sub-center ArcFace~\cite{deng2020sub} and SKH~\cite{liu2021switchable} by a large margin.

In Table.~\ref{cleandata}, we also compare our proposed ESL with other methods on the cleaned MS1MV3 dataset.
Samples in a clean class would converge to different sub-centers so that the performance of Sub-center ArcFace~\cite{deng2020sub} slightly drops.
SKH~\cite{liu2021switchable} leads to a significant performance drop when directly training on the cleaned dataset. 
The restraint of SKH~\cite{liu2021switchable} forces each hyperplane to contain a subset of all IDs in the cleaned dataset, which does great harm to the inter-class representation learning.
Compare with Sub-center ArcFace~\cite{deng2020sub} and SKH~\cite{liu2021switchable}, our proposed ESL can further boost the performance on the cleaned dataset, which further verifies the generalization of ESL.
\begin{table}[t]
    \centering
    \caption{Experiments on cleaned MS1MV3 dataset. For IJB-B and IJB-C dataset, we adopt the TPR@FPR$=$$-$$4$ as evaluation metric. }
    \resizebox{.8\textwidth}{!}{
    \begin{tabular}{l| c | c c c c c}
    \toprule[1.5pt]
        Method & Dataset & IJB-B & IJB-C & LFW & CFP-FP & AgeDB-30 \\
        \hline
        ArcFace  & MS1MV3 & 95.04 & 96.44 & $\mathbf{99.83}$ & 98.57 & 98.12  \\
        \hline
        Sub-center ArcFace M=3 & MS1MV3 & 94.84 & 96.35 & 99.75 & 98.50 & 98.14 \\
        Sub-center ArcFace M=$3\downarrow1$ & MS1MV3 & 94.87 & 96.43 & 99.78 & 98.52 & 98.19  \\
        SKH + ArcFace M=3 & MS1MV3 & 93.50  & 95.25 &  99.78 & 98.59 & 98.23  \\
        SKH + ArcFace M=$3\downarrow1$ & MS1MV3 & 94.98 & 96.48 & 99.77 & 98.70 & 98.25  \\
        \textbf{ESL + ArcFace} & MS1MV3 & $\mathbf{95.12}$ & $\mathbf{96.50}$ & 99.80 & $\mathbf{98.72}$ & $\mathbf{98.43}$  \\
    \bottomrule[1.5pt]
    \end{tabular}
    }
    \label{cleandata}
\end{table}
\\
\textbf{Generalization on Other Loss Function.}
We also verify the generalization ability of proposed ESL on CosFace~\cite{wang2018cosface}, which is another popular loss function for deep face recognition. In Tabel.~\ref{ablation_loss}, we can observe that ESL can significantly outperform Sub-center~\cite{deng2020sub} and SKH~\cite{liu2021switchable} by a large margin.

\begin{table}[t]
    \centering
    \caption{Experiments on CosFace loss function.}
    \resizebox{.8\textwidth}{!}{
    \begin{tabular}{l|c|c c c|c c c}
    \toprule[1.5pt]
        \multirow{2}{*}{Method} & \multirow{2}{*}{Dataset} &  & IJB-B &  &  & IJB-C & \\
                &         & $1e$$-$$3$ & $\mathbf{1e}$$\mathbf{-}$$\mathbf{4}$ & $1e$$-$$5$ & $1e$$-$$3$ & $\mathbf{1e}$$\mathbf{-}$$\mathbf{4}$ & $1e$$-$$5$  \\
        \hline
        CosFace                   & Mixture of Noises & 93.44 & 86.87 & 74.20 & 95.15 & 90.56 & 83.01 \\
        Sub-center CosFace M=3    & Mixture of Noises & 91.85 & 84.40 & 69.88 & 94.25 & 89.19 & 80.25 \\
        SKH + CosFace M=4         & Mixture of Noises & 95.07 & 93.15 & 87.13 & 96.28 & 94.46 & 91.87 \\
        \textbf{ESL + CosFace}             & Mixture of Noises & $\mathbf{96.52}$ & $\mathbf{94.64}$ & $\mathbf{88.93}$ & $\mathbf{97.50}$ & $\mathbf{96.10}$ & $\mathbf{93.51}$ \\
    \bottomrule[1.5pt]
    \end{tabular}
    }
    \label{ablation_loss}
\end{table}
\section{Conclusions}
In this paper, We reformulate the noise type of faces in each class into a more fine-grained manner as \noisetype{}. 
The key to robust representation learning strategies under real-world noise is the flexibility of the algorithm to the variation of \nkc.
Furthermore, we introduce a general flexible method, named Evolving Sub-centers Learning (ESL), to improve the robustness of feature representation on noisy training data. The proposed ESL enjoys scalability to different combinations of \nkc, which is more robust to unconstrained real-world noise.
Extensive experiments on noisy data training demonstrate the effectiveness of ESL and it provides a new state-of-the-art for noise-robust representation learning on large-scale noisy faces.

\setlength{\parskip}{0.5em} 
\noindent{ \bf Acknowledgments}~The work was supported by the National Key R\&D Program of China under Grant 2021ZD0201300.

%
%
\bibliographystyle{splncs04}
\bibliography{egbib}
\end{document}